\pdfoutput=1

\documentclass[11pt]{article}

\usepackage[]{emnlp2021}

\usepackage{times}
\usepackage{latexsym}

\usepackage[T1]{fontenc}

\usepackage[utf8]{inputenc}

\usepackage{microtype}

\usepackage{booktabs, multicol}
\usepackage[textsize=scriptsize]{todonotes}

\usepackage{soul}

\usepackage{enumitem}
\usepackage{booktabs}
\usepackage{multirow}
\usepackage{amsmath}
\usepackage{amsfonts}
\usepackage{amssymb}
\usepackage{fontawesome}
\usepackage[ruled,vlined]{algorithm2e}
\usepackage{arydshln}
\usepackage[procnames]{listings}
\usepackage{multicol}
\usepackage[title]{appendix}
\usepackage{graphicx} 
\usepackage{subcaption}
\usepackage{float}
\usepackage{xcolor}
\usepackage[normalem]{ulem}
\usepackage{CJKutf8}

\title{Cross-Lingual Fine-Grained Entity Typing}

\author{Nila Selvaraj, Yasumasa Onoe, Greg Durrett\\
Department of Computer Science \\
The University of Texas at Austin \\
{\tt nila.selvaraj@utexas.edu} \\ {\tt\{yasumasa, gdurrett\}@cs.utexas.edu}}

\begin{document}
\maketitle
\begin{abstract}
The growth of cross-lingual pre-trained models has enabled NLP tools to rapidly generalize to new languages. While these models have been applied to tasks involving entities, their ability to explicitly predict typological features of these entities across languages has not been established. 
In this paper, we present a unified cross-lingual fine-grained entity typing model capable of handling over 100 languages and analyze this model's ability to generalize to languages and entities unseen during training. We train this model on cross-lingual training data collected from Wikipedia hyperlinks in multiple languages (training languages). During inference, our model takes an entity mention and context in a particular language (test language, possibly not in the training languages) and predicts fine-grained types for that entity. Generalizing to new languages and unseen entities are the fundamental challenges of this entity typing setup, so we focus our evaluation on these settings and compare against simple yet powerful string match baselines. Experimental results show that our approach outperforms the baselines on unseen languages such as Japanese, Tamil, Arabic, Serbian, and Persian. In addition, our approach substantially improves performance on unseen entities (even in unseen languages) over the baselines, and human evaluation shows a strong ability to predict relevant types in these settings.

\end{abstract}

\begin{figure}[!t]
\centerline{\includegraphics[scale=.7]{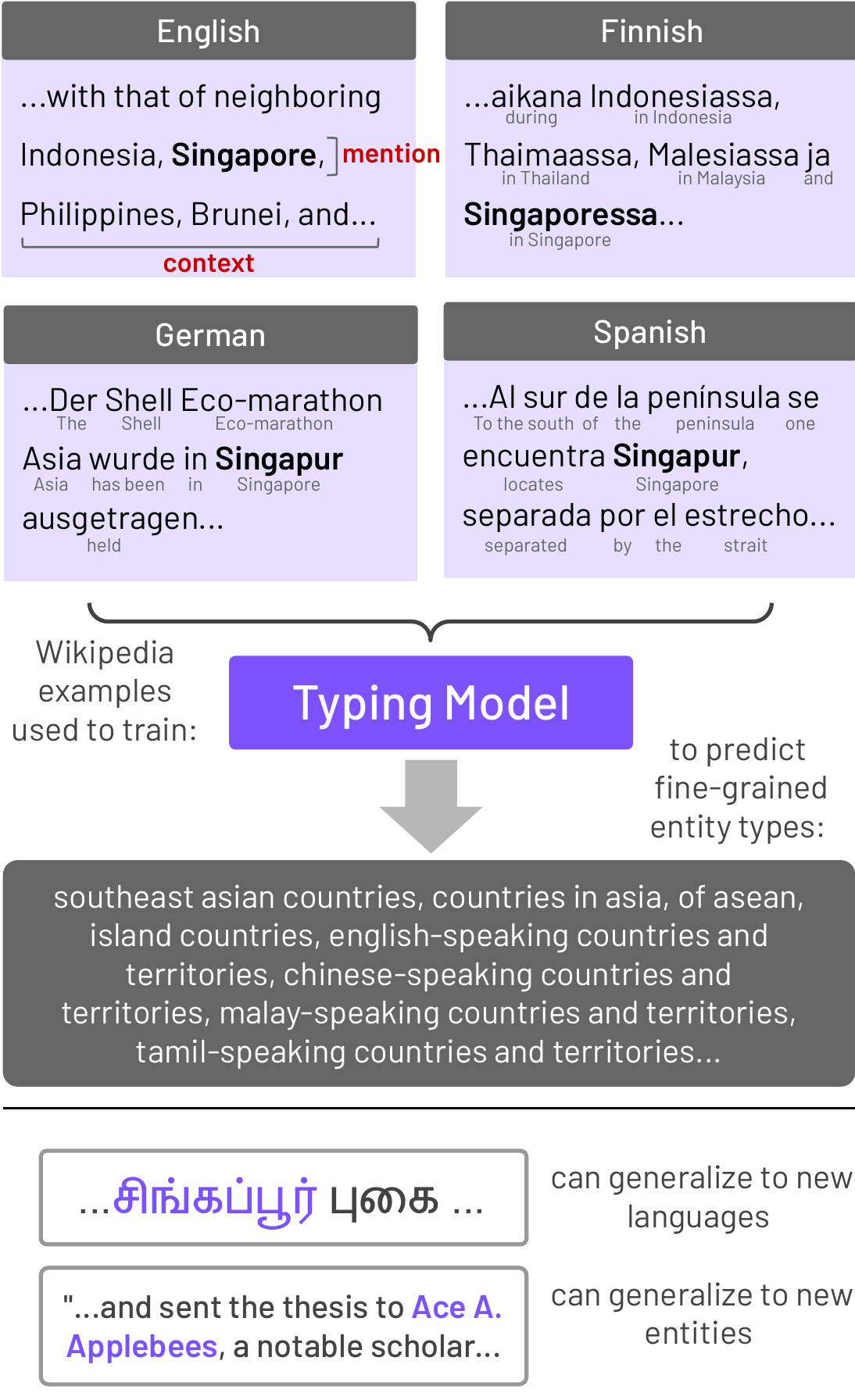}}
\caption{Cross-lingual fine-grained entity typing overview. We train a typing model using distantly supervised Wikipedia examples in English, Finnish, German, and Spanish. We evaluate the resulting model on new languages and unseen entities.}
\label{setup}
\vspace{-10pt}
\end{figure}

\section{Introduction}

Entity typing is the task of assigning types to entity mentions in natural language text \cite{bbn, Xiao_Ling_12, Dan_Gillick_14, Eunsol_Choi_18}. Fine-grained types provide richer information and are useful for many tasks, such as coreference resolution, entity linking \cite{Durrett_Klein_14, Nitish_Gupta_17, Jonathan_Raiman_18}, relation extraction \cite{Yaghoobzadeh_2017}, and question answering \cite{Lin_2012, Yavuz_2016}. Most prior work in fine-grained entity typing has predominantly focused on monolingual models in English \cite{yaghoobzadeh-schutze-2015-corpus, sonse_shimaoka_17, Dai_2019} or other often high-resource languages \cite{Van_Erp_2017, Lee_2020}. Unified cross-lingual entity typing models that can cover a wide range of languages have never been established. However, the cross-lingual setting is vital for the accessibility, inclusivity, and success of future NLP systems, to better serve more of the world's population \cite{Joshi_2020}.

In this work, we first define a cross-lingual entity typing task that associates entity mentions in any language with predefined fine-grained types without employing translation. Then, we build a unified cross-lingual entity typing model that can take an entity mention in context in over 100 different languages, outputting fine-grained types for that entity. To train this model, we automatically create entity typing datasets based on distant supervision techniques first explored in \citet{Mike_Mintz_09}, using Wikipedia articles in four languages: English, Finnish, German, and Spanish. Our typing model uses an encoder based on pre-trained multilingual BERT \cite[mBERT]{Jacob_Devlin_19} to output 10k fine-grained types derived from the Wikipedia categories. Figure~\ref{setup} shows an overview of our system.

Our evaluation primarily focuses on two questions: 1) can a cross-lingual entity typing model generalize to unseen languages? and 2) how does a cross-lingual entity typing model handle unseen entities? For the first question, we design a \emph{zero-shot languages} experimental setup in which an entity typing model is trained on the training languages (English, Finnish, German, and Spanish) and evaluated on test languages (such as Tamil) which are unseen during training. To investigate the second question, we perform experiments on \emph{unseen entities}, where the entity typing model is evaluated a set of entities held out from the model during training. 
In our experiments, we use entity typing test examples adapted from Mewsli-9 \cite{Botha_2020}, an entity linking dataset derived from WikiNews (Section~\ref{sec:test-data}). We compare our model with a string match baseline (Section \ref{sec:string-match}), which is simple but shows strong performance on the seen entities in the training languages, as well as a mention-string similarity approach. In addition, we perform human evaluation on the predicted types by our model and investigate performance breakdown by type frequency (Section~\ref{sec:human-eval} and Section~\ref{sec:analysis}).

In the \emph{zero-shot languages} setting, our approach outperforms baselines on unseen languages by a substantial margin. Surprisingly, our models can make reasonable predictions on languages with different (non-Latin) scripts from the training languages, such as Arabic, Persian, Japanese, and Tamil, which suggests that our model generalizes to unseen languages. In the \emph{unseen entities} setting, our model also shows much higher performance compared to the baselines, and can even predict plausible types for unseen entities in unseen languages. 

\section{Cross-Lingual Entity Typing}

\begin{figure*}[!t]
\centerline{\includegraphics[scale=.9]{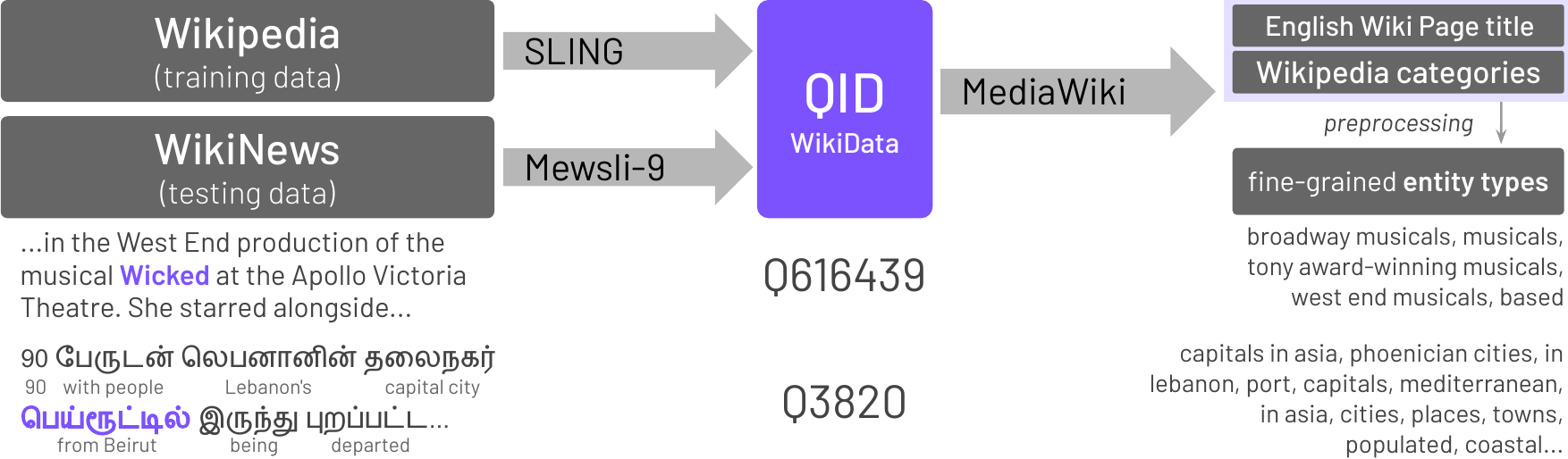}}
\caption{Data pipeline. We take the input sequence (a mention and its context) from WikiNews or WikiData, and. train a typing model using Wikipedia data from English, Finnish, German, and Spanish. We evaluate the resulting model on new languages and unseen entities.}
\label{pipeline-flowchart}
\vspace{-10pt}
\end{figure*}

Our model takes an input sequence consisting of a mention in its context. Because our model encoder is pre-trained mBERT \cite{Jacob_Devlin_19}, the model can accept input from any language of the 104 languages in the mBERT vocabulary\footnote{mBERT uses 110k subword tokens while the English BERT model has a vocabulary size of 30k; this higher number particularly includes a range of unicode characters and sequences beyond Latin script.}.
Next, a multi-label classifier outputs predictions for the input entity mention's types. The model can handle a large typeset with tens of thousands of types. We train the model using text in English, Finnish, German, and Spanish, and evaluate its performance on Arabic, German, English, Spanish, Persian, Japanese, Serbian, Tamil, and Turkish. We analyze the entity typing performance in the \emph{zero-shot languages} and \emph{unseen entities} settings (Section~\ref{sec:et-results}).

We define $ s = (w_1, ..., w_N)$ as a sequence of context words, and $m = (w_i, ..., w_j )$ as an entity mention span contained within $s$. We also define a list of types $T$ in English. Each mention and context sequence is associated with some subset of types in $T$, which we denote as $t^*$. The inputs to our cross-lingual fine-grained entity typing system consist of mention and context sequences. Given the input tuple $(m, s)$, the model $f$ outputs a vector $t \in [0,1]^{|T|}$, where each element of $t$ corresponds to one of the fine-grained types in $T$. 
Our list of predefined types $T$ is in English, but in general, the types can be in any language; they should represent language-agnostic semantic meaning.\footnote{This assumption might be flawed in two ways. First, an English typeset may be English-centric in type coverage, such as having more types around regions and states in the United States compared to other countries. Second, certain concepts in other languages may not be easily expressible in English.} By controlling the number of types or the types themselves, this entity-typing approach can generalize to different settings and knowledge bases.

\section{Data Collection}

Entity typing requires lots of annotated data to train large models, but manually annotating entities in natural text with types can be expensive, time-consuming, and tedious \cite{Eunsol_Choi_18}. This is particularly true for low-resource languages. To address this, we automatically create entity typing datasets of millions of examples from Wikipedia in several languages using distant supervision. Our data pipeline is scalable to data in new languages and automates compiling large quantities of data from Wikipedia, as seen in Figure~\ref{pipeline-flowchart}. The training data for our model consists of data which we compiled from Wikipedia in four different languages: English, German, Finnish, and Spanish. We then evaluate our model using data we adapted from Mewsli-9 \cite{Botha_2020}, a WikiNews entity linking dataset in nine different languages.

\renewcommand{\arraystretch}{1}
\begin{table}[t!]
	\centering
	\small
	\setlength{\tabcolsep}{4pt}
	\begin{tabular}{l c c c}
		\toprule
		\multicolumn{1}{l} {Language} & {Total Mentions} & {\# Sampled} & {\# Entities}\\
		\midrule
        English & 135M & \phantom{0}2.7M  & 826K \\
		Finnish & \:\:\:\:\:8M & \phantom{0}2.3M & 196K \\
		German &  \:\:\:22M & \phantom{0}2.2M & 371K \\
		Spanish &  \:\:\:40M & \phantom{0}2.7M  & 330K \\
		\midrule
		Total & $-$ & \:10.0M & $-$ \\
		Total filtered & $-$ & \phantom{0}8.9M & 1.1M \\
		\bottomrule 
	\end{tabular}
	\caption{A description of the training datasets created from Wikipedia hyperlinks, with the quantity of mentions available, number of entities, and how many we sampled for each language. }
	\label{tab:train-data}
	\vspace{-10pt}
\end{table}

\subsection{Training Data}

To train our entity typing model, we need labeled examples of $(m, s, t^*)$ triples of an entity mention, context, and gold types. We start with a dump of Wikipedia articles in four languages: English, Finnish, German, and Spanish, languages picked for their dissimilarity (among languages supported by the SLING API) and to increase entity coverage. We sample all languages in a balanced way, which in practice takes a larger fraction of data from rarer languages. Table~\ref{tab:train-data} shows a detailed breakdown of our training examples. 
We use Wikipedia hyperlinks as the mention sequences $m$, situating each mention with up to 50 context words from the article on either side of the hyperlinked text. This forms the entire context sequence $s$.

Using the SLING natural language frame semantics parser \cite{SLING},\footnote{\url{https://github.com/google/sling}} we connect each hyperlink to the WikiData QID of its target entity (the destination of the hyperlink). QIDs are language-agnostic identifiers that label each entity in the WikiData knowledge base. For each QID, we use the MediaWiki Action API to find the corresponding English Wikipedia page, and we derive the entity's gold types $t^*$ from the Wikipedia categories of this page. Lastly, we filter out examples that did not fit at least one of our 10,000 predefined types $T$ (defined in Section~\ref{sec:types}), leaving us with 8.9 million distantly-supervised entity typing training examples. 


\subsection{Test Data}\label{sec:test-data}

We derive test data from Mewsli-9 \cite{Botha_2020}, a dataset of entity mentions extracted from WikiNews articles. These entities come linked to WikiData QIDs, so we again used our pipeline with the MediaWiki Action API to annotate the mentions $m$ with gold types $t^*$ based on the categories of the corresponding English Wikipedia pages. This process gave us typing datasets in nine languages: Arabic, German, English, Persian, Japanese, Serbian, Spanish, Turkish, and Tamil. 

\subsection{Types}
\label{sec:types}

Our types are derived based on post-processing English Wikipedia categories. Using a set of rules from prior work \cite{Yasumasa_Onoe_20, Yasumasa_Onoe_20_Findings}, we map each Wikipedia category to one or more coarser types based on removing information. Specifically, we apply lowercasing, split categories on prepositions, and remove stopwords. We also split up categories with years or centuries, removing the temporal information: for example, ``20th-century atheists'' would become just ``atheists.'' These steps help reduce the frequency of highly specific types that are unlikely to be predictable from context. These post-processed Wikipedia categories become the gold types $t^*$ associated with each entity mention in our training data. Our final typeset $T$ consists of the 10,000 most frequently occurring types in the training set. See the supplementary material for some examples of common, less common, and rare types.

\section{Typing Model}


Our model $f$ accepts as input the entity mention $m$ and its context $s$ and predicts probabilities for predefined entity types $T$. Our model is similar to the one in \citet{Yasumasa_Onoe_20_Findings}, which we extend to use pre-trained multilingual BERT (mBERT) \cite{Jacob_Devlin_19}. Thus, our model accepts input in any language in the mBERT vocabulary. 

\renewcommand{\arraystretch}{1}
\begin{table*}[t]
	\centering
	\small
	\setlength{\tabcolsep}{4pt}
	\begin{tabular}{l r  r  r  r  r  r  r  r  r  r  r  r  r  r  r  r  r  r  r  r }
		\toprule
		\multicolumn{2}{c}{} & \multicolumn{3}{c}{Mention Similarity} & \multicolumn{1}{c}{} & \multicolumn{3}{c}{String Match} & \multicolumn{1}{c}{} & \multicolumn{3}{c}{English only} & \multicolumn{1}{c}{} & \multicolumn{3}{c}{Spanish only} & \multicolumn{1}{c}{} & \multicolumn{3}{c}{Multi-lang} \\
	    \cmidrule(r){3-5}  \cmidrule(r){7-9} \cmidrule(r){11-13} \cmidrule(r){15-17} \cmidrule(r){19-21}
		\multicolumn{1}{c}{Language} & \multicolumn{1}{c}{}
		 & \multicolumn{1}{c}{P} & \multicolumn{1}{c}{R} & \multicolumn{1}{c}{F1} & & \multicolumn{1}{c}{P} & \multicolumn{1}{c}{R} & \multicolumn{1}{c}{F1} & & \multicolumn{1}{c}{P} & \multicolumn{1}{c}{R} & \multicolumn{1}{c}{F1} & &
		 \multicolumn{1}{c}{P} & \multicolumn{1}{c}{R} & \multicolumn{1}{c}{F1} & &
		 \multicolumn{1}{c}{P} & \multicolumn{1}{c}{R} & \multicolumn{1}{c}{F1}\\
		\midrule
		\multicolumn{21}{c}{Unseen Languages} \\
		\midrule
		Arabic && 17.4 & 11.2 & 13.7 && 25.1 & 0.1 & 0.3 & & 61.2 & 32.6 & 42.5 & & 63.3 & 33.8 & 44.0 & & 68.4 & 33.4 & \textbf{44.9} \\
		Persian && 16.9 & 11.5 & 13.7 && 100 &  0.2 &  0.4 & &  54.4 &  34.4 &  42.1 & &  60.1 &  32.9 &  42.5 & &  58.4 &  34.1 & \textbf{43.0} \\
		Japanese &&  17.5 & 16.8 & 17.1 && 78.7 &  1.3 &  2.6 & &  43.1 &  36.6 &  39.6 & &  42.4 &  31.5 &  36.2 & &  46.7 &  36.1 & \textbf{ 40.7} \\
		Serbian &&  20.1 & 14.2 & 16.6 && 50.5 &  0.9 &  1.8 & &  69.0 &  50.7 &  58.5 & &  72.9 &  45.7 &  56.2 & &  74.1 &  58.3 & \textbf{65.3} \\
		Tamil && 12.4 & 8.2 & 9.9 && 0.0 & 0.0 & 0.0 & &  38.8 &  59.0 &  22.5 & &  37.0 &  35.0 &  19.8 & &  44.1 &  18.5 & \textbf{26.0} \\
		Turkish && 40.8 & 37.47 & 39.1 && 75.8 &  34.9 &  47.8 & &  64.9 &  55.9 &  60.1 & &  62.1 & 55.6 &  58.7 & &  68.1 &  61.1 & \textbf{64.4} \\
		\midrule
		\multicolumn{21}{c}{Seen Languages}  \\
		\midrule
		German && 55.7 & 54.8 & 55.2 && 87.1 & 74.3 & \textbf{80.2} & &  70.4 &  66.4 &  68.3 & &  72.9 &  66.4 &  69.5 & &  80.0 &  75.1 &  77.5 \\
		English &&  60.2 & 62.4 & 61.3 && 80.6 &  62.5 & \textbf{ 70.4} & &  67.3 &  63.4 &  65.3 & &  62.4 &  55.1 &  58.5 & &   68.9 &  64.1 &66.4 \\
		Spanish && 53.0 & 51.0 & 52.0 && 86.0 &  77.0 & \textbf{81.2} & &  67.8 &  63.6 &  65.6 & &  80.6 &  78.5 &  79.6 & &   78.2 &  75.8 &  77.0 \\
		\bottomrule 
	\end{tabular}
	\caption{Macro-averaged P/R/F1 on the test sets, comparing entity typing for the mention-similarity and string-match baselines, single-language models (English/Spanish), and multi-language models.} \label{tab:main-results}
	\vspace{-10pt}
\end{table*}

First, we use mBERT as the mention and context encoder. This Transformer-based encoder takes an input sequence of the form $x = \mathrm{[CLS]} m \mathrm{[SEP]} s \mathrm{[SEP]}$, with the mention $m$ and the context $s$ split into WordPiece tokens. We use the hidden vector $h^{\texttt{[CLS]}} \in \mathbb{R}^{d}$ at the $\mathrm{[CLS]}$ token as an intermediate vector representation of the mention and context, where $d$ is the dimension of hidden states. Then, we compute a dot product between $h^{\texttt{[CLS]}}$ and the type embedding matrix $\mathbf{T} \in \mathbb{R}^{d \times |T|}$ to produce a vector whose components are scores for the entity types $T$.

Finally, we pass the score vector through an element-wise sigmoid function to produce the final probabilities for each type. Each element of the vector corresponds to the model's confidence that the given input entity belongs to the corresponding type. To get the final set of predicted types for a given mention $m$ and context $s$, we add a type $k \in T$ to the set if the corresponding vector value $t_k$ is greater than a threshold value of 0.5.

Following \citet{Yasumasa_Onoe_20_Findings}, the loss is the sum of binary cross-entropy losses over all types $T$ over the whole training dataset $D$, ignoring type hierarchy in order to reduce model complexity. We predict each type independently and optimize a multi-label binary cross entropy objective:

\begin{equation*}\label{eq:loss}
\mathcal{L} = -\sum\limits_{x} \sum\limits_{k} t^*_k \cdot \log (t_k) + (1-t^*_k) \cdot \log (1- t_k)
\end{equation*}
where $t_k$ is the $k$th component of $t^*$, taking the value 1 if the $k$th type is active on the input entity mention, and $x$ represents each training example. With each iteration through the training data, we update parameters in the mBERT encoder as well as the type embedding matrix. 


\section{Experiments}

Our focus here is to shed light on the performance of our entity typing model as well as its ability to generalize along two axes: \emph{zero-shot languages} and \emph{unseen entities}. To this end, we first compare the performance of our typing model to two baselines (see Section~\ref{sec:string-match}). We also compare the performance of models trained on single language (e.g., English) with training on multiple languages. Finally, we breakdown our model's performance on unseen entities. We report macro-averaged precision, recall, and F1 metrics for our experiments.

\subsection{Baselines}\label{sec:string-match}
Given the new problem setting we tackle, few suitable baselines exist. Any pre-existing monolingual model will fail to generalize to new languages, and existing multilingual typing methods use much smaller ontologies than our model. With these limitations, we have formulated two comparison methods to contextualize our model's results.

\paragraph{String-Match Baseline}
This baseline tests how well the model can type entities by simply regurgitating types for entities it has already seen, with no disambiguation. We create a dictionary $M$ which maps all entity mention strings $m$ in the training dataset to their most frequent QID, which we then map to the corresponding gold entity types $t^*$. At test time, we predict the categories as follows:

$T = \begin{cases} M(m) &\text{if } m \text{ present in training data} \\ \emptyset &\text{otherwise }\end{cases}$

 As expected, this \textsc{String Match} approach forms a strong baseline for the test languages that are also in the training data: English, German, and Spanish. However, because this baseline only matches the exact string of an entity mention, it fails to generalize effectively to any new languages, especially those that do not use Latin characters. 
Comparing against string matching specifically highlights our system's ability to both understand the mention string itself deeply using a multilingual encoder as well as use context around the mention string to make predictions.

\paragraph{Mention String Similarity Baseline} Exact string match does not handle cases like transliteration; we implement a simple mBERT-based method to do this. For every unique entity in our training set, we encode the corresponding mention string using (non-fine-tuned) mBERT and store the hidden vector at the $\mathrm{[CLS]}$ token. At test time, we encode each example mention string in the same way, and we then perform a similarity search over the training mention representations using the FAISS library \cite{FAISS}. Finally, we predict the categories of the entity with the highest similarity to the training mention representation. 

Because we use mBERT to encode our representations, this model can make better predictions in new languages when compared to \textsc{String Match}, and it can also generalize to unseen entities if there is a semantically similar mention in the training data. However, this baseline does not have access to context.

\renewcommand{\arraystretch}{1}
\begin{table*}
	\centering
	\small
	\setlength{\tabcolsep}{4pt}
	\begin{tabular}{l r r r r r r r r r r r }
		\toprule
		\multicolumn{1}{c}{} & \multicolumn{3}{c}{Mention Similarity} & \multicolumn{1}{c}{} & \multicolumn{3}{c}{String Match} & \multicolumn{1}{c}{} & \multicolumn{3}{c}{Multi-lang} \\
	    \cmidrule(r){2-4}  \cmidrule(r){6-8} \cmidrule(r){10-12}
	    \multicolumn{1}{c}{$D'_n$}
		 & \multicolumn{1}{c}{P} & \multicolumn{1}{c}{R} & \multicolumn{1}{c}{F1} & & \multicolumn{1}{c}{P} & \multicolumn{1}{c}{R} & \multicolumn{1}{c}{F1} &  & \multicolumn{1}{c}{P} & \multicolumn{1}{c}{R} & \multicolumn{1}{c}{F1}\\
		\midrule
		Arabic$^{\prime}$ & 16.9 & 7.4 &10.3 & & 0.0 & 0.0 & 0.0 & &  64.5 &  25.4 &  \textbf{36.5} \\
		German$^{\prime}$ & 26.1 & 12.3 & 16.7 && 19.3 &  6.2 &  9.4 & &  55.0 &  31.9 &  \textbf{40.4} \\
		English$^{\prime}$ & 22.7& 15.2&18.2 && 21.3 &  6.8 &  10.3 & &  55.7 &  36.2 &  \textbf{43.9} \\
		Spanish$^{\prime}$ & 19.1 & 9.9 & 13.0 && 24.0 &  6.6 &  10.4 & &  53.3 &  28.9 &  \textbf{37.5}  \\
		Persian$^{\prime}$ & 15.5 & 5.3 & 7.9 && 85.7 &  0.4 &  0.8 & &  55.9 &  19.3 &  \textbf{28.7} \\
		Japanese$^{\prime}$ & 14.7 & 12.6 & 13.6 && 19.3 &  0.1 &  0.2 & &  44.3 &  25.8 &  \textbf{32.6} \\
		Serbian$^{\prime}$ & 19.6 & 6.3 & 9.6 && 11.1 & 0.0 & 0.0 & &  57.5 &  26.1 &  \textbf{35.9} \\
		Tamil$^{\prime}$ & 7.9 & 3.5 & 4.8 && 0.0 & 0.0 & 0.0 & &  42.8 &  14.9 &  \textbf{22.2} \\
		Turkish$^{\prime}$ & 19.0 & 9.3 & 12.5 && 29.3 &  4.5 &  7.8 & &  56.3 &  26.5 &  \textbf{36.0} \\
		\bottomrule 
	\end{tabular}
	\caption{Unseen entities. Macro-averaged P/R/F1 on the test sets, filtered to those entities which were held out from the model during training ($D'_n$), compared to the same model's performance on all the test data ($D_n$).} \label{tab:unseen-entities}
	\vspace{-10pt}
\end{table*}

\subsection{Zero-shot Languages}\label{sec:et-results}

We measure the model's performance at entity typing on six new languages: Arabic, Persian, Japanese, Serbian, Spanish, Turkish, and Tamil, evaluating how well it can generalize to new languages without any additional training data in those languages. 

\paragraph{Typing Performance}
Table~\ref{tab:main-results} reports macro-averaged precision, recall, and F1 on the test sets, compared to the \textsc{String Match} and the \textsc{Mention Similarity} baselines. Our multi-language entity typing model trained on English, Finnish, German, and Spanish data together (Multi-lang), outperforms the \textsc{Mention Similarity} baseline on all of the six test languages by a substantial margin, and even more so compared to the \textsc{String Match} baseline. We compare the Multi-lang model with models trained only on English or Spanish. Notably, our results in Table \ref{tab:main-results} show that multi-language training boosts performance on entity typing over single-language training, even in completely separate languages, with the exception of Spanish, for which the Spanish model performs slightly better. This fits with \citet{Wu_2020}, which found that high-resources languages often benefit from single-language training, but interestingly, we found that did not apply to English.

We also report performance on the training languages for completeness. The \textsc{String Match} baseline outperforms the model on these test sets in the three languages which overlap with the training set languages: English, Spanish, and German. One possible reason for this is that the entity mentions in Mewsli-9 are often unambiguous, which can be inferred from the high entity linking accuracy by the alias table baseline \cite{Botha_2020}. However, unlike the \textsc{String Match} baseline, the model is also able to generalize to new languages, even those that do not use Latin characters, so it achieves much higher results than the baseline when considering these six new languages. The model outperforms the \textsc{Mention Similarity} baseline on the all these training languages.

\begin{figure}[t]
\centerline{\includegraphics[scale=.8]{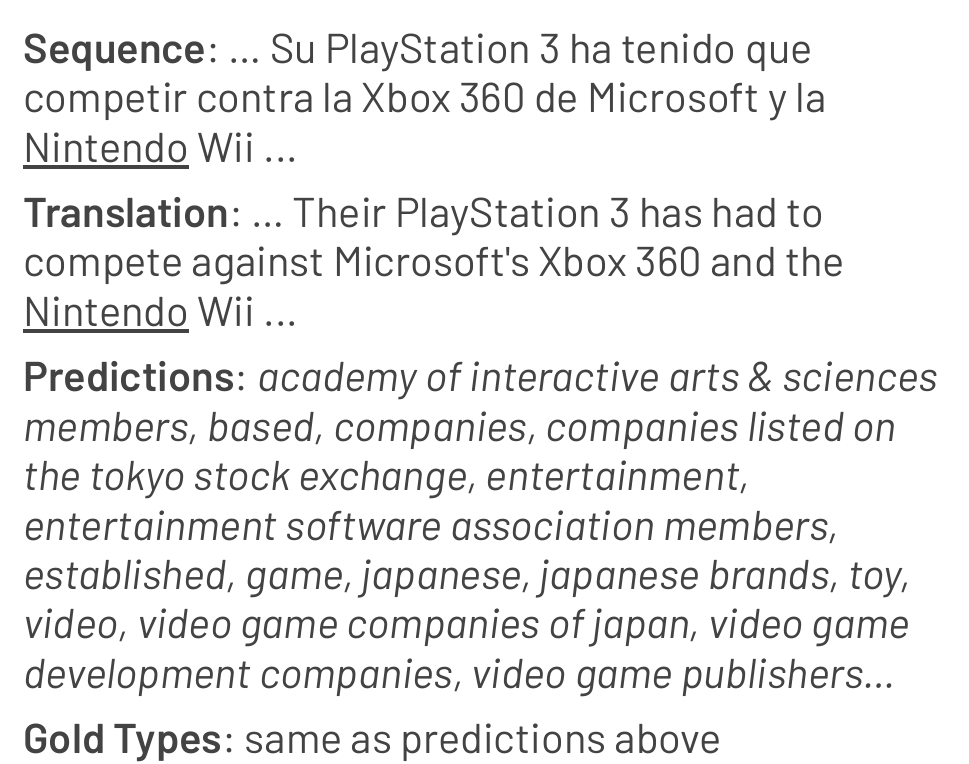}}
\caption{Model's predictions for a familiar entity in a language on which it was trained.}
\label{example1}
\vspace{-10pt}
\end{figure}

\begin{figure}[t]
\centerline{\includegraphics[scale=.8]{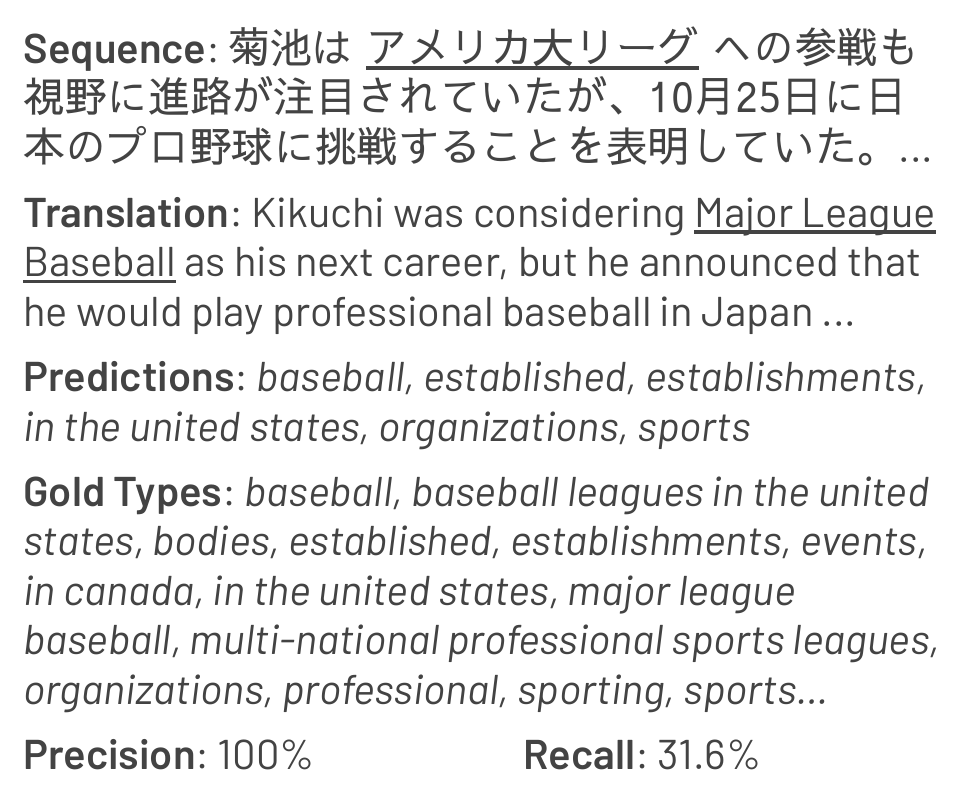}}
\caption{Model's predictions for a familiar entity in a new language.}
\label{example2}
\vspace{-10pt}
\end{figure}

\paragraph{Qualitative Analysis}
Figure~\ref{example1} shows an example of the model's predictions in Spanish, one of the languages in the training data. The model manages to predict the gold types exactly. In contrast, the example in Figure~\ref{example2} shows an input sequence in an unseen language, Japanese. Although the model does not predict every single type in the gold list for the given entity, it still picks out several of the most salient ones, and notably predicts with 100\% precision.

Our multi-language approach can be seen as an effective way to augment training data for entity typing using different languages of data available. Increasing the number of languages increases entity coverage in the training set, which boosts performance. Critically, this works even on unrelated languages. The poor performance of the \textsc{String Match} baseline on these new languages makes it evident that the entity strings do not directly match languages the model has already seen, but because types can model entities in a language-agnostic way, the model still benefits from seeing more entities during training, and it can generalize that typing knowledge to new languages. This approach may be especially useful for entity typing with low-resource languages.

\subsection{Unseen Entities}

One of the greatest difficulties in entity typing is predicting types for new entities (i.e., entities that do not appear during training), so we want to isolate these unseen types to test generalizability. In order to have a reasonable dataset to evaluate entity typing performance, we hold out certain entities from the model during training so we could later test on them. We took a random sample of 5288 entities, ensuring that we collected at least 2\% of the entities from each test set (i.e., each language). Then, we filtered out every training example referring to any of these 5288 entities to train a new model. During evaluation, we filtered the test sets to only contain examples that referred to these 5288 entities. We denote these filtered sets as $D'_n$, where $n$ represents the language.

\paragraph{Typing Performance}
Table \ref{tab:unseen-entities} shows the full results with macro-averaged precision, recall, and F1 metrics on the held-out entities as well as the original test sets. When filtering to unseen entities, the \textsc{String Match} and \textsc{Mention Similarity} baselines perform poorly, especially for languages like Arabic and Tamil. By contrast, our model is able to achieve much higher results. When predicting types for unseen entities, the model still works best on languages that overlap with the training set: English, German, and Spanish. However, it is able to predict entity types reasonably well for all languages. 

\begin{figure}[t]
\centerline{\includegraphics[scale=.8]{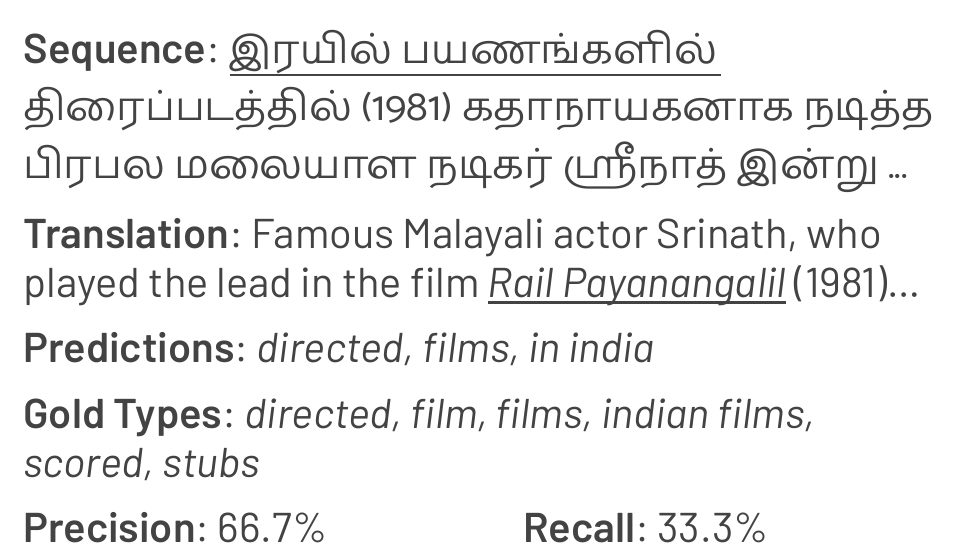}}
\caption{Model's predictions for an unseen entity in a new language.}
\label{example3}
\vspace{-10pt}
\end{figure}

\paragraph{Qualitative Analysis}
When the model's predicted types do not match the gold types, they generally still provide sensible information about the entity in question. For example, in the example in Figure~\ref{example3} with an unseen entity, the model predicts ``in india" instead of ``indian films." Some of our model's performance in Table~\ref{tab:main-results} clearly comes as a result of simply memorizing common entities in the training data; however, this memorization is not necessarily a bad thing if the model can still generalize to new languages and new entities like this one that did not appear during training.

\begin{figure}[t]
\centerline{\includegraphics[scale=.8]{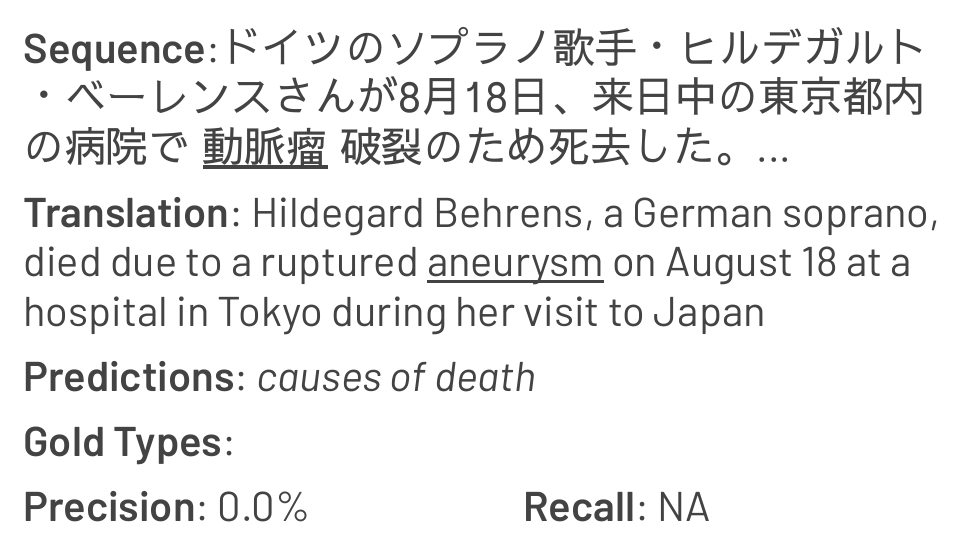}}
\caption{Model's predictions in a new language for an unseen entity which our typeset does not cover.}
\label{example4}
\vspace{-10pt}
\end{figure}

In some cases with unseen entities, our typeset is not exhaustive enough to cover an entity, so the gold types are empty, as seen in Figure~\ref{example4}. However, our model is still able to predict types that are semantically relevant. For example, in this example, ``causes of death" is a valid type when considering the mention in conjunction with its surrounding context. This example shows that despite the noisy, distantly-supervised nature of the data, the model does learn to predict types contextually.

\renewcommand{\arraystretch}{1}
\begin{table*}[t]
	\centering
	\small
	\setlength{\tabcolsep}{4pt}
	\begin{minipage}{0.69\columnwidth}
	\begin{tabular}{l r  r }
		\toprule
        \multicolumn{1}{c}{} & \multicolumn{1}{c}{Correct} & \multicolumn{1}{c}{Correct + Maybe} \\
		\midrule
		\multicolumn{3}{c}{Seen Entities ($D_n$)}\\
		\midrule
		Spanish & 91.4 & 97.6 \\
		Japanese &  83.8 &  92.8 \\
		\midrule
		\multicolumn{3}{c}{Unseen Entities ($D'_n$)}\\
		\midrule
		Spanish$^{\prime}$ & 71.4 &  89.4 \\
		Japanese$^{\prime}$ & 67.0 & 88.6  \\
		\bottomrule 
	\end{tabular}
	\caption{Macro-averaged precision for 50 examples, using human evaluation. We look at one training language (Spanish) and one zero-shot language (Japanese).} \label{tab:human-eval}
	\end{minipage}
	\hspace{16pt}
   \begin{minipage}{1.25\columnwidth}
	\begin{tabular}{l r r r r r r r r r r r}
		\toprule
		\multicolumn{1}{c}{} & \multicolumn{3}{c}{0-99} & \multicolumn{1}{c}{} & \multicolumn{3}{c}{100-999} & \multicolumn{1}{c}{} & \multicolumn{3}{c}{1000-9999} \\
	    \cmidrule(r){2-4}  \cmidrule(r){6-8} \cmidrule(r){10-12} 
		\multicolumn{1}{c}{} & \multicolumn{1}{c}{P} & \multicolumn{1}{c}{R} & \multicolumn{1}{c}{F1} & & \multicolumn{1}{c}{P} & \multicolumn{1}{c}{R} & \multicolumn{1}{c}{F1} & & \multicolumn{1}{c}{P} & \multicolumn{1}{c}{R} & \multicolumn{1}{c}{F1} \\
		\midrule
		\multicolumn{12}{c}{Seen Entities ($D_n$)}\\
		\midrule
		Arabic &  67.5 &  55.6 &  61.0 & &  39.0 &  31.7 &  34.9 & &  36.4 &  17.2 &  23.4 \\
		German &  75.6 &  88.6 &  81.6 & &  71.9 &  82.2 &  76.7 & &  73.7 &  74.7 &  74.2 \\
		English &  62.9 &  81.7 &  71.1 & &  57.9 &  69.1 &  63.0 & &  59.4 &  59.2 &  59.3 \\
		Spanish &  70.9 &  87.3 &  78.2 & &  68.9 &  80.4 &  74.2 & &  66.9 &  72.5 &  69.6 \\
		Persian &  63.5 &  61.9 &  62.7 & &  38.8 &  29.9 &  33.8 & &  27.8 &  16.9 &  21.0 \\
		Japanese &  39.8 &  59.6 &  47.7 & &  40.3 &  40.9 &  40.6 & &  27.0 &  26.0 &  26.5 \\
		Serbian &  76.0 &  73.3 &  74.6 & &  61.8 &  55.4 &  58.4 & &  62.2 &  52.1 &  56.7 \\
		Tamil &   46.6 &  35.5 &  40.3 & &  23.0 &  14.2 &  17.6 & &  12.0 &  6.8 &  8.7 \\
		Turkish &   68.1 &  76.1 &  71.9 & &  61.7 &  63.2 &  62.4 & &  51.6 &  51.9 &  51.8 \\ 
		\midrule
		\multicolumn{12}{c}{Unseen Entities ($D'_n$)}\\
		\midrule
        Arabic$^{\prime}$ &  67.9 &  50.7 &  58.1 & &  26.7 &  15.4 &  19.5 & &  13.1 &  6.1 &  8.3 \\
		German$^{\prime}$ &  60.9 &  48.2 &  53.8 & &  42.6 &  27.7 &  33.6 & &  34.4 &  19.0 &  24.5  \\
		English$^{\prime}$ &  57.2 &  57.2 &  57.2 & &  45.2 &  37.3 &  40.8 & &  37.6 &  22.7 &  28.3\\
		Spanish$^{\prime}$ &  55.1 &  46.2 &  50.3 & &  38.9 &  30.5 &  34.2 & &  31.1 &  21.3 &  25.3 \\
		Persian$^{\prime}$ &  61.1 &  41.4 &  49.4 & &  27.6 &  13.1 &  17.7 & &  14.0 &  4.7 &  7.1 \\
		Japanese$^{\prime}$ &  34.9 &  38.4 &  36.6 & &  31.4 &  25.8 &  28.3 & &  23.5 &  22.0 &  22.7 \\
		Serbian$^{\prime}$ &  74.9 &  50.9 &   60.6 & &  44.7 &  10.6 &  17.1 & &  7.1 &  2.4 &  3.6 \\
		Tamil$^{\prime}$ &  34.2 &  44.3 &  25.9 & &  16.3 & 6.9 & 9.7 & & 4.3 & 2.3 & 0.3 \\
		Turkish$^{\prime}$ &  60.4 &  42.0 &  49.5 & &  42.2 &  22.9 &  29.7 & &  23.8 &  11.9 &  15.9 \\ 
		\bottomrule 
	\end{tabular}
	\caption{Macro-averaged P/R/F1 on the test sets, broken down into buckets by the types' frequency rankings. Results are shown both for the full dataset ($D_n$) and on unseen entities ($D'_n$).}\label{tab:type-frequency}
	\vspace{-10pt}
	\end{minipage}
\end{table*}


\subsection{Human Evaluation on Predicted Types}\label{sec:human-eval}

Due to the distantly-supervised nature of our testing data, we may be underestimating the precision; as seen in the previous examples, some types that our model predicts may be relevant even if they do not exist in the actual Wikipedia categories. We take one seen language (Spanish) and one unseen language (Japanese) as case studies to perform human evaluation. We examine the model's predicted types which did not exist in the gold types for 50 examples with seen entities and 50 examples with unseen entities in each language. For each type such that $t \notin t^{*}$, we assign a label of either \emph{correct}, \emph{incorrect}, or \emph{maybe}. We use the third category to describe types where either the meaning or scope of the type is unclear, such as ``autonomous," or those that are closely related but may narrowly fail to apply to this particular entity, such as predicting ``pandemics'' for the entity ``HIV.''

We report our results in Table~\ref{tab:human-eval}. After including the correct types by human evaluation, macro-averaged precision increased significantly in both Spanish and Japanese, increasing even further when also including types with the \emph{maybe} labels. The model performs poorly with certain kinds of entities with lots of highly specific type information, especially those that cannot typically be inferred by context. For example, with many sports figures, the model predicts a wide variety of semi-related categories about their origins, teams, or positions, many of which are factually incorrect and sometimes mutually exclusive, such as predicting both ``association football forwards'' and ``association football midfielders.'' However, even when the model predicts a type outside the set of gold types, the majority of these predictions are still correct. Many of these examples highlight the model's strengths in predicting types using contextual information, which the Wikipedia categories could not make use of and which is not inherently ``taught'' by our training data. 


\subsection{Rare Types}\label{sec:analysis}

We finally break down entity typing performance by the type frequency ranking in three buckets: [0, 99], [100, 999], and [1000, 9999]. We report macro-averaged precision, recall, and F1 metrics in Table~\ref{tab:type-frequency}, both on the full dataset and on unseen entities. For the languages in the training data (English, German, Spanish), entity-typing performance is relatively high across all three buckets. The model also achieves good results on unseen languages for frequent types. Most of the categories in this first bucket are more coarse-grained, corresponding to certain locations or professions, and the model performs relatively well at distinguishing these types. 

Next, we look at the type-frequency breakdown of entity typing limited to unseen entities. For predicting the rarest types on unseen entities, performance drops off in the second and third buckets for certain languages, such as Tamil, Serbian, Arabic, and Persian. However, the model is able to predict rare types on unseen entities in other unseen languages, such as Japanese and Turkish. Although the numbers here are hard to draw strong conclusions from, given the results of the previous human evaluation, they show some success even on the most challenging cases.

\section{Related Work}

\paragraph{Entity Typing}
Entity typing has been a long-studied subtask in NLP, characterized by a growth in the size and complexity of type sets, from 4 \cite{tjong-kim-sang-de-meulder-2003-introduction} to 17 \cite{Eduard_Hovy_06} to hundreds and even thousands \cite{Eunsol_Choi_18, Xiao_Ling_12, Dan_Gillick_14}. Recent work has shifted towards predicting fine-grained entity types types because these types offer much richer information and are more useful for various downstream NLP tasks, such as entity linking, coreference resolution, and question answering \cite{Durrett_Klein_14, Lin_2012, Yavuz_2016}. 

\paragraph{Cross-Lingual Models}

The task of cross-lingual entity typing has not yet been explored much. Most prior works focus on one languages at a time, such as English  \cite{yaghoobzadeh-schutze-2015-corpus}, Japanese \cite{Suzuki_2016}, Dutch and Spanish \cite{Van_Erp_2017}, or Chinese \cite{Lee_2020}. Additionally, these models predominantly focus on fewer than one hundred types. For example, \citet{Nothman_2013} performs named entity recognition on nine different languages, but they only use coarse-grained types. 

\paragraph{Other Multilingual Tasks}
Pretrained contextual representation models trained on masked language modeling \cite{Jacob_Devlin_19, Conneau_2020} have set a new standard for many NLP systems. They have achieved surprising success in multilingual settings even without explicit cross-lingual signals by pre-training on text from multiple languages with one unified vocabulary. This growth has motivated work in cross-lingual modeling for tasks such as natural language inference, document classification, named entity recognition, part-of-speech tagging, and dependency parsing  \cite{Wu_2019, Pires_2019}. However, further analysis has indicated a high level of variety in their performance across different languages and tasks \cite{Hu_2020}. \citet{Wu_2020} observed that BERT does not learn high-quality representations particularly for low-resource languages, suggesting that these languages require more data or more efficient pre-training techniques. 

\paragraph{Knowledge Probing and Knowledge in BERT} Our model's capability to memorize tracts of Wikipedia connects to past work on extracting knowledge from language models \cite{Petroni_2019}; our task is easier because it is also possible to make type predictions based on context, giving our model a ``backoff'' capability. Recently, knowledge probing settings have been proposed for the multilingual setting as well \cite{Kassner_2021}. Another related line of work attempts to add entity information into BERT \cite[inter alia]{Zhang_2019,Peters_2019,Poerner_2020,Fevry_2020}; our model could potentially benefit from this, but given our focus on generalizing to new entities, a model that more easily memorizes common entities may actually work less well.

\section{Conclusion}
In this work, we have presented an approach to fine-grained cross-lingual entity typing. Our typing model accepts input in over 100 different languages and outputs categories in English. We generated a dataset of distantly supervised Wikipedia typing examples in four different languages, and we show that the multi-language model outperformed single-language models, even on completely new languages. We also show that this model was able to generalize to completely new languages, predicting rare types on unseen entities, and using contextual information to do so.

\bibliography{anthology,custom}
\bibliographystyle{acl_natbib}

\clearpage

\appendix

\section*{Appendix A: Typeset}
The following lists show the top 10 types in our three frequency groups. We use frequency as an approximation for sorting our types from course-grained to fine-grained. The coarse-grained types are useful to infer distinctions like living vs. dead people (\textit{living people} vs \textit{deaths}). They also highlight locations with differing granularities (continents, countries, cities) and some relational information (like \textit{in europe}). The rarer categories become highly specific and fine-grained types. 
\paragraph{Coarse-Grained}
\begin{enumerate}[nosep]
  \item established
  \item establishments
  \item births
  \item people
  \item deaths
  \item states
  \item territories
  \item living people
  \item places
  \item in europe
\end{enumerate}

\paragraph{Fine-Grained}
\begin{enumerate}[nosep]
\setcounter{enumi}{100}
  \item province
  \item of finland
  \item government
  \item players
  \item personnel
  \item competitions
  \item geography
  \item politicians
  \item film
  \item unit
\end{enumerate}

\paragraph{Very Fine-Grained}
\begin{enumerate}[nosep]
\setcounter{enumi}{1000}
  \item republican
  \item former capitals of the united states
  \item of the community of portuguese language countries
  \item in the roman empire
  \item former portuguese colonies
  \item in new spain
  \item british islands
  \item emigrants
  \item association football
  \item of scotland
\end{enumerate}

\begin{figure}[t]
\centerline{\includegraphics[scale=.78]{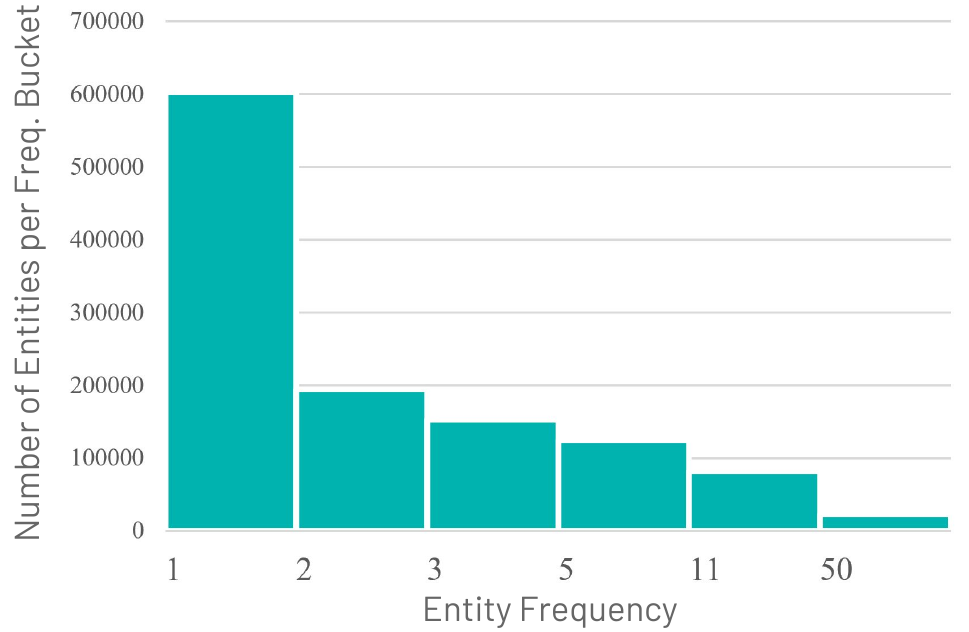}}
\caption{Distribution of entities in the training set, showing the number of entities in each frequency bucket.}
\label{entity-dist-train}
\end{figure}

\section*{Appendix B: Selected Examples}
\label{sec:appendix-entity-pred}

We present some example predictions in Figure~\ref{entity-predictions} below. The model can make accurate predictions on unseen entities, such as ``DMDK,'' ``TerraSAR-X,'' and ``Spirited Away,'' especially on coarser-grained types. Even when our set of gold types is empty according to Wikipedia categories, the model can use contextual information to output relevant types, as in the case of ``openSUSE'' or ``magnitude.'' For common entities such as countries, the model typically memorizes the gold types and can apply this knowledge even to entity mentions in new languages, like Tamil. However, sometimes this memorization leads the model to fail to account for contextual information, such as with the ``Paris" example below.  

\begin{figure*}[]
\centerline{\includegraphics[scale=.77]{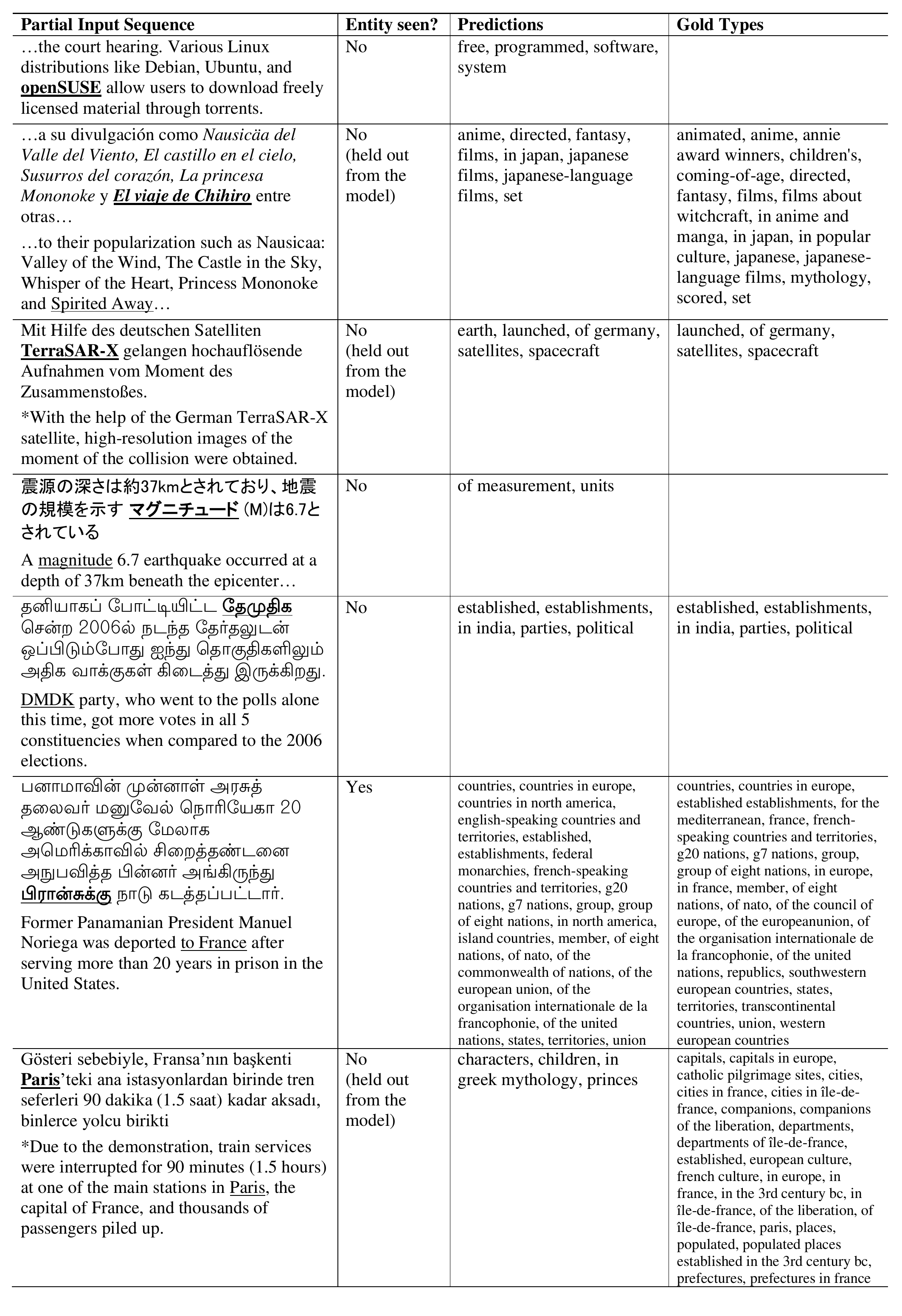}}
\caption{Selected examples of typing predictions across different languages. Starred translations were obtained using Google Translate and may be inaccurate. }
\label{entity-predictions}
\end{figure*}.

\section*{Appendix C: Distribution of Entities}
\label{sec:appendix-entity-dist}
Figure~\ref{entity-dist-train} shows how entities are distributed across the training set. 

\section*{Appendix D: Hyperparameters}
\label{sec:appendix-hyperparam}
\paragraph{Hyperparameters} We use pre-trained mBERT-base-uncased \cite{Jacob_Devlin_19} for our mention and context encoder. We train our model with batch size 16 using NVIDIA V100 GPUs. We use the AdamW optimizer \cite{Kingma_14, Ilya_Loshchilov_18} with learning rate 2e-5 for BERT parameters and learning rate 1e-3 for the type embedding matrix. We use Pytorch \cite{Pytorch} and the HuggingFace Transformers library \cite{Thomas_Wolf_19} to implement our models.

\section*{Appendix E: Limitations of the Data Collection Pipeline}

Our data collection approach has a few limitations. We picked English Wikipedia categories because English Wikipedia has the most comprehensive category annotations. However, we had to omit some language-specific entities that did not have a corresponding English Wikipedia page, which account for about 20\% of the non-English hyperlinks. Additionally, Wikipedia articles only hyperlink the first instance of an entity mention in an article, which often skew towards less ambiguous entity mention strings. For example, the first mention of a person typically includes their full name, while subsequent mentions often only contain their last name.

Furthermore, it would be interesting to test our approach on a domain entirely outside Wikipedia or Wikidata. However, we focus specifically on fine-grained entity typing, and no suitable datasets exist already, as annotating datasets with large and fine-grained ontologies is very challenging.

\section*{Appendix F: Testing Datasets}
Table~\ref{tab:test-data} provides the number of examples in the testing datasets derived from Mewsli-9 \cite{Botha_2020}.

\renewcommand{\arraystretch}{1}
\begin{table}[t]
	\centering
	\small
	\setlength{\tabcolsep}{4pt}
	\begin{tabular}{l c c c}
		\toprule
		\cmidrule(r){2-2} \cmidrule(r){4-4} 
		\multicolumn{1}{c}{Language} & \# Examples ($D_n$) & & \# Examples ($D'_n$)\\
		\midrule
		Arabic & \:\:7038 && \:\:1995 \\
		German & 60631 && 13661 \\
		English & 79675 && 10589 \\
		Spanish & 47064 && 10878 \\
		Persian & \:\:\:\:521 && \:\:\:\:127 \\
		Japanese & 27731 && \:\:4662  \\
		Serbian & 34944 && 20414 \\
		Tamil & \:\:2660 && \:\:\:\:938\\
		Turkish & \:\:5530 && \:\:1430 \\
		\bottomrule 
	\end{tabular}
	\caption{A description of the test datasets derived from Mewsli-9. $D_n$ 
	refers to the total test size of each dataset per language. $D'_n$ refers to test datasets filtered to examples with entities we held out during training in order to test the model's ability to generalize to new languages. }
	\label{tab:test-data}
\end{table}

\end{document}